\documentclass{article} 
\usepackage{iclr2023_conference,times}

\usepackage{amsmath,amsfonts,bm}

\def\eqref#1{equation~\ref{#1}}

\def\1{\bm{1}}

\DeclareMathAlphabet{\mathsfit}{\encodingdefault}{\sfdefault}{m}{sl}
\SetMathAlphabet{\mathsfit}{bold}{\encodingdefault}{\sfdefault}{bx}{n}

\usepackage{ragged2e}
\usepackage{hyperref}
\usepackage{url}

\title{How good are Commercial Large Language Models on African Languages?}

\author{Jessica Ojo  \\
Masakhane\\
\texttt{jessicaojo19@gmail.com} \\
\And
Kelechi Ogueji \\
Masakhane\\
\texttt{kelechi.ogueji@uwaterloo.ca} \\
}

\iclrfinalcopy 
\begin{document}

\maketitle

\begin{abstract}

Recent advancements in Natural Language Processing (NLP) has led to the proliferation of large pretrained language models.
These models have been shown to yield good performance, using in-context learning, even on unseen tasks and languages.
They have also been exposed as commercial APIs as a form of language-model-as-a-service, with great adoption.
However, their performance on African languages is largely unknown.
We present a preliminary analysis of commercial large language models on two tasks (machine translation and text classification) across eight African languages, spanning different language families and geographical areas.
Our results suggest that commercial language models produce below-par performance on African languages.
We also find that they perform better on text classification than machine translation.
In general, our findings present a call-to-action to ensure African languages are well represented in commercial large language models, given their growing popularity.

\end{abstract}

\section{Introduction}
\label{intro}

Large language models have risen to the fore of Natural Language Processing (NLP).
These models have been shown to achieve state-of-the-art performances on several tasks.
More recently, focus has shifted from the pretrain-finetune paradigm \citep{howard-ruder-2018-universal,devlin-etal-2019-bert,roberta2019,2020t5} to in-context learning \citep{NEURIPS2020_1457c0d6,xglm,wei2022finetuned,palm,scaling-iflm,sanh2022multitask,in-context-survey}.
In-context learning proves that prompting large language models with some task-specific examples allows them perform well on test examples of that task, all without updating the model's parameters.
This has led to reduced computation costs and has made it possible to create language-models-as-a-service \citep{sun2022bbt}, in the form of commercial Application Programming Interfaces (APIs).
Commercial language models have become very prevalent.
For context, the recently released ChatGPT\footnote{{\url{https://chat.openai.com/}}} amassed 100 million users\footnote{{\url{https://www.theguardian.com/technology/2023/feb/02/chatgpt-100-million-users-open-ai-fastest-growing-app}}} in two months, making it the fastest growing consumer app in recent history.
Given their dominance and inevitable continual rise, it is important to understand how these models perform on African languages.
Hence, we present a preliminary effort to close this gap by evaluating two commercial large language models using in-context learning on African languages.
Evaluation is performed on two tasks - text classification and machine translation.
Our experiments, spanning 8 African languages from different language families and geographical locations, suggests that commercial language models do not perform well on African languages.
In particular, we note a large disparity in performance, depending on the evaluation task - models perform better on text classification than machine translation.
Our work sheds light on the need to ensure the inclusion of African languages in the development of commercial language models, given their inevitable adoption in our daily lives.

\section{Related Work}
\label{related}

\subsection{In-Context Learning}
The use of pretrained language models has become the de-facto approach to solving natural language processing (NLP) tasks.
Previous models such as BERT \citep{devlin-etal-2019-bert}, RoBERTa \citep{roberta2019} and T5 \citep{2020t5} largely follow a pretrain-finetune setting \citep{howard-ruder-2018-universal}.
In this method, the pretrained model is finetuned on a downstream task, such as text classification, and then used for that task.
While this works very well, it has several downsides.
For one, finetuned models are usually task-specific and this means one has to maintain separate models for separate tasks.
Furthermore, the growing size of pretrained language models \citep{scaling-laws} means that it is becoming increasingly expensive to finetune such gigantic models.
One solution that has proven popular in recent times is in-context learning \citep{NEURIPS2020_1457c0d6,schick-schutze-2021-just,wei2022finetuned,palm,scaling-iflm,sanh2022multitask,in-context-survey}.
The core idea behind this method is to enable pretrained language models learn from examples within the context.
In this setting, a user prompts a pretrained language model with a few labelled examples of a task following a specific pattern, and unlabelled examples that need to be predicted on \citep{wei2022chain,liu-etal-2022-makes,wei2022emergent}.
In-context learning can also work in a zero-shot setting where no labelled examples are included in the prompt.
In-context learning works surprisingly well and is very efficient since there is no update to the model's parameters.
As a result, computation costs are significantly reduced and it becomes possible to expose language models as a service \citep{sun2022bbt}.
Commercial APIs are heavily reliant on in-context learning as this is the primary method through which users interact \footnote{{\url{https://platform.openai.com/docs/guides/completion/prompt-design}}} with the models\footnote{{\url{https://docs.cohere.ai/docs/prompt-engineering}}}.

\subsection{Multilingual In-Context Learning}
Large language models have proven successful in multilingual settings.
\citet{xglm} train several multilingual models, of which the largest one (7.5B parameters) sets a state-of-the-art in few-shot learning on more than 20 languages.
Their model outperforms GPT3 on several multilingual tasks.
\citet{crosslingual-multitask} perform multitask prompted finetuning on multilingual pretrained language models and observe impressive zero-shot generalization to tasks in unseen languages.
Following findings from \cite{language-contamination} that non-English dataset present in the pretraining corpora of English language models explains their surprising cross-lingual ability, \citet{palm} deliberately introduce non-English corpora ($\approx 22\%$) into the pretraining corpora of their PaLM model and achieve impressive few-shot multilingual performance.
\citet{wei-multilingual} evaluate GPT3 and PaLM on a newly introduced grade school mathematics multilingual benchmark.
They find that using prompts with intermediate reasoning steps in English consistently led to competitive or better results than those written in the native language of the question.
They also set a new state-of-the-art on a common-sense reasononing multilingual benchmark, XCOPA \citep{ponti-etal-2020-xcopa}, using few-shot examples.
\citet{zhao-schutze-2021-discrete} show that prompting yields better cross-lingual transfer in few-shot settings than finetuning and in-language training of multilingual natural language inference.
Furthermore, \citet{winata-etal-2021-language} evaluate the multilingual ability of GPT \citep{radford2019language} and T5 \citep{2020t5} models on multi-class text classification, and find that they work well on non-English languages given a few English examples.
Concurrent work \citep{chatpgpt-translator} evaluate ChatGPT on machine translation and find that, while it is competitive with other commercial translation APIs such as Google translate\footnote{{\url{https://translate.google.com/}}}, it is less robust on other domains such as biomedical.
Another concurrent work \citep{llm-translator} conducts a study on the performance of GLM \citep{zeng2022glm130b} on machine translation.
They note several interesting findings on the effect of prompt template, examples and language.
Despite the plethora of works on multilingual prompting, little to no African languages are usually contained in the evaluation sets of nearly all of these works.
When present, they are often obtained by translating the existing datasets of other languages \citep{yu2022counting}
This method has been shown to contain artifacts that can inflate the performance of models evaluated on such datasets \citep{artetxe-etal-2020-translation}.
Our work is orthogonal to all of this works because we focus solely on commercial language model APIs, given their prevalence.
The closest to our work is concurrent by \citet{lelapa2023}, who evaluate GPT 3.5 on Named Entity Recognition and Machine Translation on only isiZulu.
However, our work is different from this as we compare two commercial APIs in the evaluation of text classification and Machine Translation across 8 African language.

\section{Methodology}
\label{method}

\subsection{Datasets}
Evaluation is done on two tasks - text classification and machine translation.

\subsubsection{Text Classification}
We use the news topic classification datasets from \citet{hedderich-etal-2020-transfer} and \citet{alabi-etal-2022-adapting}.
We select the Hausa (hau) language from \citet{hedderich-etal-2020-transfer} which has 5 categories.
Pretrained language models have been shown to work very well on this dataset in both few and zero-shot settings.
The dataset from \citet{alabi-etal-2022-adapting} covers five languages, out of which we select four - Nigerian Pidgin (pcm), Malagasay (mlg), and Somali (som), isiZulu (zul).
Each language has 5 categories, except Somali which has 6.
For both datasets, we use the train, validation and test splits as released by the authors.
We select these languages because they cover different language families and geographical areas.

\subsubsection{Machine Translation}
We use the MAFAND-MT machine translation dataset from \citet{adelani-etal-2022-thousand} which covers 16 African languages.
Running translation on commercial APIs is cumbersome and expensive, hence we select 5 languages from the 16. 
The five languages are isiZulu (zul), Yoruba (yor), Nigerian Pidgin (pcm), Swahili (Swa) and Lugala (lug).
We use the splits as released by the authors.

\subsection{Models}
Two commercial APIs\footnote{Experiments were run between January 22, 2023 and February 5, 2023.} are considered: ChatGPT\footnote{{\url{https://chat.openai.com/}}} and Cohere\footnote{{\url{https://www.cohere.ai}}}.
We consider both these APIs because they are arguably the most popular ones\footnote{{\url{https://venturebeat.com/uncategorized/openai-rival-cohere-launches-language-model-api/}}}.
ChatGPT is based on the Instruct-GPT models \citep{ouyang2022training}.
It is optimized for conversations and has been shown to be capable of several NLP tasks including text classification, machine translation, question answering, and so on.
We use Cohere's multilingual model\footnote{{\url{https://docs.cohere.ai/changelog/multilingual-support-for-coclassify}}} which is based on their multilingual embedding model\footnote{{\url{https://docs.cohere.ai/docs/multilingual-language-models}}}.
The embedding model supports 100 languages, including 15 African languages.
All the languages we consider, except Nigerian Pidgin, are supported by the model.
However, given the linguistic proximity of Nigerian Pidgin to English \citep{faraclas2008nigerian,ogueji-ahia-pidginunmt,pidgin_pivot,ahia-ogueji-towards,lent-etal-2021-language,lent-etal-2022-creole}, the model should be able to perform well on the dataset. 

\subsection{Prompting and Evaluation}
We describe our prompting and evaluation approaches for text classification and machine translation.

\subsubsection{Text Classification}
For Cohere, we use the Classify\footnote{{\url{https://api.cohere.ai/classify}}} endpoint and follow the format specified in the API documentation\footnote{{\url{https://docs.cohere.ai/reference/classify}}}.
When using ChatGPT, we design several prompts ourselves and we also ask ChatGPT for the best prompt for classification, following concurrent work \citep{chatpgpt-translator}.
We perform some initial evaluation of the prompts and select the best one.

Our best prompt is shown below:
\begin{quote}
\small
\texttt{Given the following news headlines and their categories:}\\
\texttt{Text: \{Sentence\}}\\
\texttt{Category: \{Label\}}\\

\texttt{Please classify the following news headlines into one of: \{Label List\}.}\\
\texttt{Text: \{Sentence\}}\\\
\texttt{Category: }\\
\end{quote}
Where $Sentence$ is the news headline to be classified, $Category$ is the news topic, and $Label List$ is a comma separated list of all unique labels for that language.

For both models, we supply two example demonstrations per category from the training set.
We randomly sample 100 samples from the test set for each language and evaluate on this.
Both demonstrations and evaluation are done across two random seeds, such that we sample distinct demonstrations and test samples for each language with each random seed.
We report the average F1 score for each language across both seeds.
It should be noted that we decide to evaluate on a subset of the test set because of the tedious nature of obtaining results ChatGPT.

\subsubsection{Machine Translation}
We do not use Cohere for machine translation because its generation API currently supports only English\footnote{{\url{https://docs.cohere.ai/docs/generation-card\#technical-notes}}}.
ChatGPT is used for all our machine translation evaluations.
Preliminary results from comparing few-shot to zero-shot translations on Nigerian Pidgin suggested no noticeable difference.
Hence, we perform all translations in a zero-shot manner because of the tedious nature and low-throughput of obtaining results from ChatGPT.

We use the prompt used in concurrent work \citep{chatpgpt-translator} which is shown below:

\begin{quote}
\small
\texttt{Please provide the [TGT] translation for these sentences:}\\
\texttt{\{Sentence\}}\\
\texttt{\{Sentence\}}\\
\end{quote}

Where $TGT$ is the target language to be translated into, and $Sentence$ is a sentence to be translated.
We sample 100 sentences from the test set of each language and evaluate translating this to and from English. 
We report the BLEU score \citep{papineni-etal-2002-bleu} which is calculated using SacreBLEU \citep{post-2018-call}.

It has been shown that English prompts perform better, on average, than in-language prompts \citep{xglm,wei-multilingual}, so we do not explore prompting in the target language for both tasks.

\section{Results}
\label{results}

\subsection{Text Classification}
Results are reported in table \ref{class_results}.
As we can see, both commercial models fall well below the current state of the art.
Surprisingly, Cohere's multilingual embedding model is the worst performer, despite supporting almost all the languages evaluated on.
Nigerian Pidgin has the highest score in the Cohere results.
This is most likely as a result of its close linguistic relationship with English language, which usually makes up a significant portion of the pretraining corpora of pretrained language models \citep{wenzek-etal-2020-ccnet,pile,roots}.
ChatGPT is the best performing commercial model, and it gets above average F1 scores on all languages.
Similar to Cohere, Hausa and Nigerian Pidgin possess the highest F1 scores.
The details of ChatGPT's pretraining corpora and exact training methods are unknown, so it is hard to hypothesize a reason for its relatively good performance.
However, it is very likely that its pretraining corpora contains non-English text.
Furthermore, multilinguality has been shown to be a part of possible emergent abilities of large language models \citep{wei2022emergent}, so the performance is not entirely surprising.
Overall, both commercial models fall significantly short of the current state of the art.
While ChatGPT is the better performer, Cohere's performance is especially surprising since it has been trained on almost all of the evaluated languages\footnote{{\url{https://txt.cohere.ai/multilingual/}}}.

\begin{table}[t]
\caption{Text Classification Results: We report the F1 scores for the commercial models. We also report the current state of the art result obtained from \citet{alabi-etal-2022-adapting}. Best results per language are in bold.}
\label{class_results}
\begin{center}
\begin{tabular}{llll}
\multicolumn{1}{c}{\bf Language}  &\multicolumn{1}{c}{\bf Cohere} &\multicolumn{1}{c}{\bf ChatGPT} &\multicolumn{1}{c}{\bf Current SOTA}
\\ \hline \\
Hausa ($hau$)             &43.2 & 77.9& \textbf{91.2}\\
Malagasay ($mlg$)             &35.0 & 51.1& \textbf{67.3}\\
Nigerian Pidgin ($pcm$)         & 48.8  & 73.4 & \textbf{82.2} \\
Somali ($som$)             & 28.4 & 51.3 & \textbf{79.9}\\
isiZulu ($zul$)             & 24.8 & 54.8 & \textbf{79.6}\\
\end{tabular}
\end{center}
\end{table}

\subsection{Machine Translation}
Results are reported in table \ref{mt_results}.
ChatGPT has very poor performance on machine translation, obtaining BLEU scores of less than 1.0 on all languages.
This is very surprising given its good performance on text classification.
Our results agree with concurrent work \citep{lelapa2023} which finds that GPT 3.5 obtains a BLEU score of 0 on Zulu to English translation. Our findings are also somewhat similar to \citep{chatpgpt-translator}, which reports significantly worse performance on Romanian, a relatively low-resource language, than on higher-resource languages like English and German.
While the BLEU scores are too low to draw conclusions from, ChatGPT seems to perform better when translating into English than from it.
This agrees with previous works \citep{belinkov-etal-2017-neural,bugliarello-etal-2020-easier} which show that it is harder to translate into morphologically rich languages, like African ones, than morphologically poor ones like English.
In general, our results suggest that ChatGPT is not good enough for translation involving African languages.
It also suggests that ChatGPT performs better on sequence classification tasks than it does on text generation tasks for African languages.

\begin{table}[t]
\caption{Machine Translation Results: We report the BLEU scores of the translations from ChatGPT. We also report the current state of the art result obtained from \citet{adelani-etal-2022-thousand}} and \citet{nllb2022}. Best results per language are in bold.
\label{mt_results}
\begin{center}
\begin{tabular}{lll}
\multicolumn{1}{c}{\bf Translation Direction}  &\multicolumn{1}{c}{\bf ChatGPT} &\multicolumn{1}{c}{\bf Current SOTA}
\\ \hline \\
Lug$\rightarrow$Eng         & 0.16 & \textbf{30.9}  \\
Eng$\rightarrow$Lug             & 0.13 & \textbf{25.8} \\
\\ \hline \\
Pcm$\rightarrow$Eng         & 0.22 & \textbf{45.2} \\
Eng$\rightarrow$Pcm             & 0.20 & \textbf{35.0} \\
\\ \hline \\
Swa$\rightarrow$Eng         &0.18  & \textbf{39.3} \\
Eng$\rightarrow$Swa             &0.15  & \textbf{30.7}\\
\\ \hline \\
Yor$\rightarrow$Eng         & 0.10 & \textbf{24.4}\\
Eng$\rightarrow$Yor             & 0.12 & \textbf{14.4}\\
\\ \hline \\
Zul$\rightarrow$Eng         & 0.31 & \textbf{40.3} \\
Eng$\rightarrow$Zul             & 0.26 & \textbf{22.9} \\

\end{tabular}
\end{center}
\end{table}

\section{Error Analysis}
We take a closer look at some errors made by the model on machine translation. Specifically, we focus on two languages - Yoruba and Nigerian Pidgin - because they are understood by the authors. For each language, we randomly select 3 samples and discuss their predictions.

\subsection{Yoruba translations}
Samples are shown in table \ref{yor_analysis}.
Looking at sample 1, ChatGPT mistranslates ``B{\'i} omi b{\'a} gb{\'o}n{\'a} ju b{\'i} {\'o} {\d s}e y{\d e} l{\d o}" which means ``When water becomes too hot" to ``Water is poured into the container".
Furthermore, the English to Yoruba translation is completely wrong and riddled with a lot of misspellings and grammatical errors.
In sample 3, ChatGPT gets the translations wrong and also transposes the words ``obìnrin" (woman) and ``okùnrin" (man) in the translations.
One notable observation across English to Yoruba translations is that ChatGPT does not always include diacritics in its Yoruba predictions.
Overall, ChatGPT does a really poor job in translating in either direction.
The hallucinatory nature of the model predictions is evident, as all translations barely have any correlation with the original sentences.

\begin{table}[t]
\caption{Examples of Nigerian Pidgin translation using ChatGPT}
\label{pidgin_analysis}
\begin{center}
\begin{tabular}{p{0.8in} p{1.3in} p{1.3in} p{1.3in}}
\multicolumn{1}{c}{\bf }  &\multicolumn{1}{c}{\bf Sample 1 }  &\multicolumn{1}{c}{\bf Sample 2 }  &\multicolumn{1}{c}{\bf Sample 3}
\\ \hline \\
\RaggedRight{Nigerian Pidgin Sentence}        &{\small Commission head, Simon Byabakama, assure di nation on top live TV for Thursday evening afta polls don close say result don dey enter for di national tally centre despite di nationwide internet blackout} &{\small Di goal be give Eyimba some hope as dem kontinu to press for plunty nomba to secure a point at least} &{\small Na im inauguartion be di first time wey elected leaders dey handova give each oda for Niger history for over 60 years wey dem don get independence, but di attack from jihadist and some kind plan to do coup don spoil di celebrate.}\\
\\ \hline \\
\RaggedRight{English Sentence}    &{\small Commission head, Simon Byabakama, assured the nation on live TV on Thursday evening after polls closed that results were arriving at the national tally centre despite the nationwide internet blackout.} &{\small The goal gave Enyimba some hope as they continued to press in numbers to at least secure a point} &{\small His inauguration marks the first handover between elected leaders in Niger's 60-year post-independence history, but jihadist attacks and an alleged attempted coup have overshadowed ceremonies.}\\
\\ \hline \\
\RaggedRight{ChatGPT Translation Pcm$\rightarrow$Eng}   &{\small Commission head, Simon Byabakama, assured the nation on live TV on Thursday evening after the polls had closed that the results were entering the national tally center despite the nationwide internet blackout.} &{\small The goal was to give Eyimba some hope as they continue to press for a sufficient number of goals to secure at least one point.} &{\small The inauguration marked the first time in over 60 years of independence that elected leaders have handed power to each other in Niger's history, but attacks by jihadists and plans for a coup have disrupted the celebrations}\\
\\ \hline \\
\RaggedRight{ChatGPT Translation Eng$\rightarrow$Pcm }  &{\small Simon Byabakama wey dey head the Commission, e talk for TV on Thursday night after election close say the result don dey reach the centre where dem dey count the result even though internet all over the place dey cut.} &{\small The goal give Enyimba some hope as dem still dey press wey full force to fit get at least one point.} &{\small The day wey dem swear am for office, na the first time wey leader wey get vote go hand over to another leader since 60 years after independence for Niger but the attacks wey the bad people wey dey carry gun dey do, and the way wey dem try put eye for the power no gree am to happen make e no dey sweet for the day.}\\

\end{tabular}
\end{center}
\end{table}

\begin{table}[t]
\caption{Examples of Yoruba translation using ChatGPT}
\label{yor_analysis}
\begin{center}
\begin{tabular}{p{0.8in} p{1.3in} p{1.3in} p{1.3in}}
\multicolumn{1}{c}{\bf }  &\multicolumn{1}{c}{\bf Sample 1 }  &\multicolumn{1}{c}{\bf Sample 2 }  &\multicolumn{1}{c}{\bf Sample 3}
\\ \hline \\
\RaggedRight{Yoruba Sentence}        & {\small B{\'i} omi b{\'a} gb{\'o}n{\'a} ju b{\'i} {\'o} {\d s}e y{\d e} l{\d o} (t{\`a}b{\'i} b{\'i} {\'o} b{\'a} tut{\`u} j{\`u}) {\`a}w{\d o}n iy{\`u}n y{\'o}{\`o} l{\`e} ew{\`e} omi — t{\'i} y{\'o}{\`o} s{\'i} p{\`a}d{\'a}n{\`u} {\d o}w{\d o} t{\`o} {\'o} {\'n} fi o{\`u}nj{\d e} n{\`u} {\'u}n.} & {\small K{\'i} {\'o} t{\'o} de {\`e}{\'e}r{\'o}o {\`a}g{\d o} n{\'a}{\`a}, erin o{\d s}{\`u} m{\d e}{\d e}f{\`a} n{\'a}{\`a} ti lu ok{\`u}n.} & {\small {\`I}k{\d e}l{\`u} t{\'i} {\'o} k{\d o} lu {\`a}w{\d o}n ob{\`i}nrin w{\d o}ny{\'i} ({\`a}ti {\`a}w{\d o}n {\d o}k{\`u}nrin kan) ror{\`o} b{\'i} {\d e}ranko {\d e}h{\`a}nn{\`a}.}\\
\\ \hline \\
\RaggedRight{English Sentence}    &{\small When water becomes too hot (or too cold) the corals expel the algae — and with it, its main source of food.} &{\small Before his arrival at the camp, the six-month-old was snared in a wire trap.}  &{\small The attacks on these women (and some men) are vicious and inhuman.}\\
\\ \hline \\
\RaggedRight{ChatGPT Translation Yor$\rightarrow$Eng}   &{\small Water is poured into the container (or it is filled) until the container is full - when it overflows, the excess water is poured out.} &{\small If you do not pay attention to your work, the work of others will be affected.} &{\small The tradition requires men to cover their heads (and sometimes women too) in public places.}\\
\\ \hline \\
\RaggedRight{ChatGPT Translation Eng$\rightarrow$Yor }  &{\small Ti omi ba n mu ki i f{\d e}{\d e} (tabi ki i f{\d e}{\d e}) aw{\d o}n corals ni w{\d o}n fi omi omi p{\d o} - ati ni it{\d o}ju r{\d e}, iru ib{\d e}r{\d e} ti a n pa.} &{\small Bayi o ti gba l{\d o} si ibeere, {\d o}m{\d o} t{\d o}l{\d o}ti si igba ni a ti dara ju {\d o}nu.} &{\small {\`I}d{\'a}m{\`o} g{\d e}g{\d e} b{\'i} i {\`a}w{\d o} n {\d o}m{\d o}-{\`o}m{\`o}-w{\'e} ({\`a}ti n{\'i} k{\`o}k{\`o}r{\`o}) n{\`i}y{\'i}n ni {\`a}j{\`i}n{\`a} {\`i}w{\'e}j{\`u} {\`i}t{\d o}l{\d o}m{\d o} w{\'a}j{\`u}.}\\

\end{tabular}
\end{center}
\end{table}

\subsection{Nigerian Pidgin Translations}
Samples are shown in table \ref{pidgin_analysis}.
Looking at the Nigerian Pidgin sentences, we can see the language's linguistic similarity with English.
Interestingly, while the ChatGPT predictions yield low BLEU scores, they are somewhat semantically similar to the ground truth.
However, there notable errors made across board.
For example, focusing on the Nigerian Pidgin to English predictions in sample 2, there are tense errors.
Also, the model seems to misunderstand what ``numbers" refers to in the input text, as its prediction indicates it confuses it for the number of goals.
Furthermore, across all samples, the model seems to be poor at translating certain English words to Nigerian Pidgin words, such as ``The" to ``Di", so it always retains the original English word.
In general, while the predictions in both directions for all samples have notable issues, they are more semantically similar to the ground truth than the BLEU scores suggests. 
This highlights the drawbacks of automatic metrics based on N-gram overlap.


\section{Conclusion}
\label{conclusion}
We have presented a preliminary analysis of commercial language models on African languages.
\citet{joshi-etal-2020-state} note that over 90\% of the world’s 7000+ languages are under-studied by the NLP community.
Despite the 2000+ spoken languages and over 1 billion people in Africa\footnote{{\url{https://en.wikipedia.org/wiki/Demographics_of_Africa}}}, its languages make up a significant portion of the under-studied languages \citep{blasi-etal-2022-systematic}.
While there have been several efforts \citep{nekoto2020participatory,towards-dblp,masakhaner,ogueji-etal-2021-small,nllb2022,alabi-etal-2022-adapting,dossou2022afrolm,serengeti} to close this gap, there is still a lot of work to be done.
This is even more pertinent given the rapid commercial adoption of large scale language models.
Our findings suggest that these models do not perform well on African languages. 
In particular, there seems to be performance disparity, depending on the task evaluated.
Although our work reports what is, to the best of our knowledge, the first evaluation of commercial language models on African languages, we note that this only a preliminary study that needs to be further advanced.
Future works could focus on more advanced prompting methods such as chain-of-thought \citep{wei2022chain} and pivot prompting \citep{chatpgpt-translator}, evaluation of more test samples and a wider variety of tasks.
While our finding may be impacted by the sampled test data, the use of the BLEU automatic metric \citep{callison-burch-etal-2006-evaluating,mathur-etal-2020-tangled,freitag-etal-2020-bleu}, prompting template and prompting examples, it nonetheless presents a call-to-action to ensure African languages are well represented in the age of commercial large language models.


\bibliography{iclr2023_conference}
\bibliographystyle{iclr2023_conference}


\end{document}